# Parametric Return Density Estimation for Reinforcement Learning

**Tetsuro Morimura**
IBM Research - Tokyo
tetsuro@jp.ibm.com

**Masashi Sugiyama**
Tokyo Institute of Technology
sugi@cs.titech.ac.jp

**Hisashi Kashima**
The University of Tokyo
kashima@mist.i.u-tokyo.ac.jp

**Hirotaka Hachiya**
Tokyo Institute of Technology
hachiya@sg.cs.titech.ac.jp

**Toshiyuki Tanaka**
Kyoto University
tt@i.kyoto-u.ac.jp

## Abstract

Most conventional Reinforcement Learning (RL) algorithms aim to optimize decision-making rules in terms of the expected returns. However, especially for risk management purposes, other risk-sensitive criteria such as the value-at-risk or the expected shortfall are sometimes preferred in real applications. Here, we describe a parametric method for estimating *density* of the returns, which allows us to handle various criteria in a unified manner. We first extend the Bellman equation for the conditional expected return to cover a conditional probability density of the returns. Then we derive an extension of the TD-learning algorithm for estimating the return densities in an unknown environment. As test instances, several parametric density estimation algorithms are presented for the Gaussian, Laplace, and skewed Laplace distributions. We show that these algorithms lead to risk-sensitive as well as robust RL paradigms through numerical experiments.

## 1  Introduction

Most reinforcement learning (RL) methods attempt to find decision-making rules that maximize the *expected* return, where the return is defined as the cumulative total of (discounted) immediate rewards. Most of the theories in RL have been developed for working with the expected return as the objective function. However, users are sometimes interested in seeking big wins on rare occasions or avoiding a small chance of suffering a large loss. Actually, risk control is one of the central topics in financial engineering. For example, in portfolio theory, analysts are expected to find a portfolio that maximizes profit while suppressing the risks of large losses (Luenberger, 1998).

Since maximizing the expected return does not necessarily avoid rare occurrences of large negative outcomes, we need other criteria for risk evaluation. Recently, such risk-averse scenarios have also been considered in RL as *risk-sensitive RL* (Heger, 1994; Mihatsch and Neuneier, 2002; Bagnell, 2004). Well-written reviews are found in Geibel and Wysotzki (2005) and Defourny et al. (2008).

In this paper, we describe an approach to handling various risk-sensitive and/or robust criteria in a unified manner, where a return density is estimated and then these criteria are evaluated from the estimated return density. In order to achieve this purpose, we first show that the Bellman equation, which is a recursive formula for the (conditional) expectation of the returns, can be extended to a (conditional) *probability density* of the returns in Section 3. The extended Bellman equation gives the return density as its solution. However, it has functional degrees of freedom as a drawback in practice. To address this problem, in Section 4, we next describe a parametric approach on the basis of the stochastic natural gradient method as an extension of the temporal-difference (TD) learning. Several parametric density estimation algorithms for the Gaussian, Laplace, and skewed Laplace distributions are presented. We show that the algorithm with the Gaussian distribution can be regarded as a natural extension of the conventional TD learning, and the algorithm for the Laplace distribution leads to robust RL. As an example of using the estimated return density, in Section 5, we focus on a quantile of the return, also known as the value-at-risk in finance, and provide practical RL algorithms. In Section 6, numerical experiments show that the proposed algorithms are promising in risk-sensitive and robust RL scenarios.

## 2  Background of Value-based RL

We briefly review the framework of value-based RL and our motivation to estimate return densities.

## 2.1 Markov Decision Process (MDP)

An RL problem is usually defined on a discrete-time Markov decision process (MDP) (Bertsekas, 1995; Sutton and Barto, 1998). It is defined by the quadruplet $(\mathcal{S}, \mathcal{A}, p_{\mathrm{T}}, p_{\mathrm{R}})$, where $\mathcal{S} \ni s$ and $\mathcal{A} \ni a$ are sets of states and actions, respectively. The state transition probability $p_{\mathrm{T}} : \mathcal{S} \times \mathcal{A} \times \mathcal{S} \to [0,1]$ is a function of a state $s$, an action $a$, and the successor state $s_{+1}$, as $p_{\mathrm{T}}(s_{+1}|s,a) \triangleq \Pr(s_{+1}|s,a)$. We describe the state $s_{+k}$ and the action $a_{+k}$ as a state and an action after $k$ time-steps from $s$ and $a$, respectively. An immediate reward $r \in \mathbb{R}$ is randomly generated according to the reward density $p_{\mathrm{R}} : \mathbb{R} \times \mathcal{S} \times \mathcal{A} \times \mathcal{S} \to \mathbb{R}^+$, which is a function of $r$, $s$, $a$, and $s_{+1}$, as $p_{\mathrm{R}}(r|s,a,s_{+1}) \triangleq \frac{d}{dr}\Pr(R \leq r|s,a,s_{+1})$. Here, we consider a (stationary) policy $\pi : \mathcal{A} \times \mathcal{S} \to [0,1]$ as a decision-making rule of an learning agent, which is a probability function of $a$ given $s$, such as $\pi(a|s) \triangleq \Pr(a|s)$.

## 2.2 From expected return to return density

Let us define the return that is the time-discounted cumulative reward with a discount rate $\gamma \in [0,1)$,

$$\eta \triangleq \lim_{K \to \infty} \sum_{k=0}^{K} \gamma^k r_{+k}, \tag{1}$$

where $r_{+0}$ is identical to $r$. This is observed by the learning agent with (infinite) time delay and usually is a random variable $E$, reflecting randomness due to $p_{\mathrm{T}}$, $p_{\mathrm{R}}$, and $\pi$. Once the policy is fixed, the MDP is regarded as a Markov chain $\mathrm{M}(\pi) \triangleq \{\mathcal{S}, \mathcal{A}, p_{\mathrm{T}}, p_{\mathrm{R}}, \pi\}$. We write the return probability density functions as

$$p_{\mathrm{E}}^\pi(\eta \mid s,a) \triangleq \frac{d}{d\eta} \Pr(E \leq \eta \mid s, a, \mathrm{M}(\pi)).$$

While various statistics of the returns could be used for criteria of the objective or constraint, in ordinary value-based RL, the expected return is often used for the objective function such as

$$Q^\pi(s,a) \triangleq \mathbb{E}^\pi\{\eta \mid s,a\},$$

where $\mathbb{E}^\pi\{\cdot\}$ denotes the expectation over $\mathrm{M}(\pi)$. $Q^\pi(s,a)$ is often called the state-action value or Q function. However, decision-making with relying only on information on the expected return will be insufficient for the control of the risk. Furthermore, the expected return is not robust against outliers, *i.e.*, its estimate can be severely affected by rare occurrences of huge noises which could be included in the reward- or state-observations (Sugiyama et al., 2009).

Altogether, a major drawback of the ordinary RL approach follows the fact that the approach omitted all information on the returns except for the expectations. In actuality, if we have an approximation for the return density, it allows us to access a lot of information on the returns and to handle various risk-sensitivity and/or robustness criteria for the returns. This is why we focus on the return density estimation.

## 2.3 Related work on return density

There are several studies to estimate features of the return density. In the seminal paper of Dearden et al. (1998), a recursive formula for the second moment of the return is derived for a deterministic reward case as $\mathbb{E}^\pi\{\eta^2|s\} = r^2 + 2\gamma V^{\pi,\gamma}(s_{+1}) + \gamma^2 \mathbb{E}^\pi\{\eta^2|s_{+1}\}$. They developed a Bayesian learning method with the normal-gamma distribution. However, their approach requires numerical integrations, while our proposed algorithm does not require any numerical integration. As a similar line to Dearden et al. (1998), several *mean-variance* model-free RL algorithms are developed in Sato et al. (2001) and Engel et al. (2005), and were successful in the variance penalized MDPs. However, these algorithms assume that the return density is Gaussian, which may not be true in practice. In contrast, our approach can choose the density model for the return. As a similar work to this paper, Morimura et al. (2010) propose a nonparametric approach to estimating the return distribution.

## 3 Distributional Bellman Equation

Here we derive a Bellman-type recursive formula for the return density, which we term the *distributional* Bellman equation for the return density[1].

**Proposition 1** *The distributional Bellman equation for the (conditional) return density is given by*

$$\begin{aligned} p_{\mathrm{E}}^\pi(\eta|s,a) &= \frac{1}{\gamma} \sum_{s_{+1} \in \mathcal{S}} p_{\mathrm{T}}(s_{+1}|s,a) \int_r p_{\mathrm{R}}(r|s,a,s_{+1}) \\ &\quad \times \sum_{a_{+1} \in \mathcal{A}} \pi(a_{+1}|s_{+1}) p_{\mathrm{E}}^\pi\left(\frac{\eta-r}{\gamma}\Big|s_{+1},a_{+1}\right) dr \\ &\triangleq \Pi^\pi p_{\mathrm{E}}^\pi(\eta|s,a), \end{aligned} \tag{2}$$

*where $\Pi$ is the distributional Bellman operator.*

**Proof Sketch:** From the definition of the return in Eq. (1), its recursive form is $\eta = r + \gamma \eta_{+1}$. The random variables $R$ and $E_{+1}$ are conditionally independent given the successor state $s_{+1}$ since these are distributed as $r \sim p_{\mathrm{R}}(r|s,a,s_{+1})$ and $\eta_{+1} \sim p_{\mathrm{E}}^\pi(\eta_{+1}|s_{+1})$, respectively. Given a return value $\eta$ and a reward value $r$, the return $\eta_{+1}$ has to be $\eta_{+1} = (\eta-r)/\gamma$. Thus,

$$\begin{aligned} \Pr(E \leq \eta \mid s,a) &= \sum_{s_{+1} \in \mathcal{S}} p_{\mathrm{T}}(s_{+1}|s,a) \int_r p_{\mathrm{R}}(r|s,a,s_{+1}) \\ &\quad \times \sum_{a_{+1} \in \mathcal{A}} \pi(a_{+1}|s_{+1}) \Pr\left(E_{+1} \leq \frac{\eta-r}{\gamma} \mid s_{+1}, a_{+1}\right) dr, \end{aligned}$$

---
[1] The distributional Bellman equation in Morimura et al. (2010) is for the cumulative density, not for the density.

holds. Eq. (2) is obtained by differentiating the above equation with respect to $\eta$. □

## 4 Approximation of Return Density with Parametric Model

The distributional Bellman equation shown in Proposition 1 gives the return density as its solution. However, it will be hard to deal with the distributional Bellman equation in practice due to its functional degrees of freedom. We propose a principled method to circumvent this difficulty, by introducing a parametric family $\mathcal{P}$ to estimate the return density.

### 4.1 Return density approximation with distributional Bellman equation

We consider adjusting an estimate $\hat{p}_\mathrm{E}$ close to $\Pi^\pi \hat{p}_\mathrm{E}$ for $\hat{p}_\mathrm{E} \in \mathcal{P}$ in terms of a divergence between $\hat{p}_\mathrm{E}$ and $\Pi^\pi \hat{p}_\mathrm{E}$, and solving it with gradient descent. For this purpose, we use the Kullback–Leibler (KL) divergence from $\Pi^\pi \hat{p}_\mathrm{E}$ to $\hat{p}_\mathrm{E}$ such as, for a state-action pair $(s, a)$,

$$D_\mathrm{KL}[\Pi^\pi \hat{p}_\mathrm{E}(\eta|s,a), \hat{p}_\mathrm{E}(\eta|s,a)]$$
$$\triangleq \int_\eta \Pi^\pi \hat{p}_\mathrm{E}(\eta|s,a) \log \frac{\Pi^\pi \hat{p}_\mathrm{E}(\eta|s,a)}{\hat{p}_\mathrm{E}(\eta|s,a)} \mathrm{d}\eta. \quad (3)$$

It is noted that this approach follows the same line as the *expectation propagation* (Minka, 2001) and has a "moment matching" property[2] if the applied parametric distribution $p_\mathrm{E}^\pi$ belongs to an exponential family.

### 4.2 Parametric models for the return density approximation

Let us consider the cases where the return density is modeled by the Gaussian, Laplace, or skewed Laplace distribution.

(i) The Gaussian distribution with parameter $\boldsymbol{\theta}^\mathrm{g} \triangleq [\mu, \sigma]^\top$ is 'normal' and symmetric, where $\top$ denotes transpose:

$$p^\mathrm{g}(x \mid \mu, \sigma) \triangleq \frac{1}{\sqrt{2\pi}\sigma} \exp\left(-\frac{1}{2\sigma^2}(x-\mu)^2\right), \quad (4)$$

where $\mu \in \mathbb{R}$ and $\sigma > 0$ are the mean and standard deviation parameters, respectively. Note that this belongs to the exponential family.

(ii) The Laplace distribution with parameter $\boldsymbol{\theta}^\mathrm{l} \triangleq [m, b]^\top$ is symmetric and fat-tailed:

$$p^\mathrm{l}(x \mid m, b) \triangleq \frac{1}{2b} \exp\left(-\frac{1}{b}|x-m|\right), \quad (5)$$

---
[2]The moment matching means that moments of an estimated density $q_x$ in the optimal solution match these of the targeted density $p_x$ (Bishop, 2006).

where $m \in \mathbb{R}$ and $b > 0$ are the central and scale parameters, respectively. Note that $m$ is equal to the 0.5-quantile (*i.e.*, the median) and the arithmetic mean.

(iii) The skewed Laplace distribution with parameter $\boldsymbol{\theta}^\mathrm{skl} \triangleq [m, b, c]^\top$ is asymmetric and fat-tailed:

$$p^\mathrm{skl}(x \mid m, b, c)$$
$$\triangleq \frac{c(1-c)}{b} \begin{cases} \exp\left(\frac{1-c}{b}(x-m)\right) & \text{if } x < m, \\ \exp\left(-\frac{c}{b}(x-m)\right) & \text{otherwise,} \end{cases} \quad (6)$$

where $m \in \mathbb{R}$, $b > 0$, and $c \in (0, 1)$ are the central, scale, and skewness parameters, respectively. Note that $m$ is equal to the $c$-quantile.

We chose the above models since they allow us to obtain (stochastic) gradients analytically for the KL divergence and also analytic estimators of the $q$-quantile return, as shown in Sections 4.3 and 5, respectively.

It is noted that every conditional return density given each possible state-action pair is required in the RL scenario. Thus we suppose that the parameter in the return density model is determined as a function of the state-action pair, *e.g.*, $\boldsymbol{\theta}^\mathrm{g}(s, a) = [\mu_{\boldsymbol{\theta}}(s, a), \sigma_{\boldsymbol{\theta}}(s, a)]^\top$, where the subscript $\boldsymbol{\theta}$ in $\mu_{\boldsymbol{\theta}}$ and $\sigma_{\boldsymbol{\theta}}$ is a (total) parameter of the return density model.

### 4.3 (Natural) gradient descent approach

We consider a gradient of the KL divergence in Eq. (3) for a return density model $\hat{p}_\mathrm{E}(\eta|s, a, \boldsymbol{\theta})$ parametrized by a parameter $\boldsymbol{\theta} \in \mathbb{R}^d$. To be consistent with the approach of the conventional TD learning (Sutton and Barto, 1998), both the parameters for $p_\mathrm{E}^\pi$ and $\Pi^\pi \hat{p}_\mathrm{E}$ ought to be treated as different parameters $\boldsymbol{\theta}$ and $\boldsymbol{\theta}'$, respectively, while $\boldsymbol{\theta}'$ is usually set to $\boldsymbol{\theta}$, *i.e.*,

$$\frac{\partial}{\partial \theta_i} D_\mathrm{KL}[\Pi^\pi \hat{p}_\mathrm{E}(\eta|s,a,\boldsymbol{\theta}'), \hat{p}_\mathrm{E}(\eta|s,a,\boldsymbol{\theta})]\Big|_{\boldsymbol{\theta}'=\boldsymbol{\theta}}$$
$$= \lim_{\varepsilon \to 0} \{D_\mathrm{KL}[\Pi^\pi \hat{p}_\mathrm{E}(\eta|s,a,\boldsymbol{\theta}), \hat{p}_\mathrm{E}(\eta|s,a,\boldsymbol{\theta}+\varepsilon \boldsymbol{e}_i)]$$
$$\qquad - D_\mathrm{KL}[\Pi^\pi \hat{p}_\mathrm{E}(\eta|s,a,\boldsymbol{\theta}), \hat{p}_\mathrm{E}(\eta|s,a,\boldsymbol{\theta})]\}/\varepsilon$$
$$= -\int_\eta \Pi^\pi \hat{p}_\mathrm{E}(\eta|s,a,\boldsymbol{\theta}) \frac{\partial}{\partial \theta_i} \log \hat{p}_\mathrm{E}(\eta|s,a,\boldsymbol{\theta}) \mathrm{d}\eta, \quad (7)$$

where the vector $\boldsymbol{e}_i$ is a $d$-dimensional vector of zeros except that the $i$-th element is 1.

The gradient given in Eq. (7) leads to the following gradient descent update rule with a notation for a function $f(\boldsymbol{\theta})$, $\partial f(\boldsymbol{\theta})/\partial \boldsymbol{\theta} \triangleq [\partial f(\boldsymbol{\theta})/\partial \theta_1, \ldots, \partial f(\boldsymbol{\theta})/\partial \theta_d]^\top$,

$$\boldsymbol{\theta} := \boldsymbol{\theta} + \alpha \int_\eta \Pi^\pi \hat{p}_\mathrm{E}(\eta|s,a,\boldsymbol{\theta}) \frac{\partial}{\partial \boldsymbol{\theta}} \log \hat{p}_\mathrm{E}(\eta|s,a,\boldsymbol{\theta}) d\eta,$$

where $:=$ denotes the right-to-left substitution and $\alpha$ is a sufficiently small and positive learning rate.

However, since a parametric density family $\mathcal{P}_{\boldsymbol{\theta}} \triangleq \{\hat{p}_{\mathrm{E}}(\eta|s,a,\boldsymbol{\theta}) \,|\, \boldsymbol{\theta} \in \mathbb{R}^d\}$ will form a manifold structure with respect to the parameter $\boldsymbol{\theta}$ under the KL divergence, instead of a Euclidean structure, the ordinary gradient obtained in Eq. (7) does not properly reflect the differences in the sensitivities and the correlations between the elements of $\boldsymbol{\theta}$ for the probability densities $\hat{p}_{\mathrm{E}}(\eta|s,a,\boldsymbol{\theta})$. Because of this, the ordinary gradient is generally different from the steepest direction on the manifold and the optimization process often becomes unstable or falls into a stagnant state, called a *plateau* (Amari et al., 2000; Kakade, 2002).

To avoid the problem of plateaus, the *natural gradient* (NG) method was proposed by Amari (1998), which is a gradient method on a Riemannian space. A parameter space is a Riemannian space if a parameter $\boldsymbol{\theta} \in \mathbb{R}^d$ is on a Riemannian manifold defined by a positive definite matrix called a Riemannian metric matrix $\boldsymbol{F}(\boldsymbol{\theta}) \in \mathbb{R}^{d \times d}$. The squared length of a small incremental vector $\Delta\boldsymbol{\theta}$ connecting $\boldsymbol{\theta}$ to $\boldsymbol{\theta} + \Delta\boldsymbol{\theta}$ in a Riemannian space is given by $\|\Delta\boldsymbol{\theta}\|_{\boldsymbol{F}}^2 = \Delta\boldsymbol{\theta}^\top \boldsymbol{F}(\boldsymbol{\theta})\Delta\boldsymbol{\theta}$. Since we use the KL divergence for measuring divergence between densities, an appropriate Riemannian metric has to inherit characteristics from the KL divergence. In this situation, the Fisher information matrix (FIM) of $\mathcal{P}_{\boldsymbol{\theta}} \ni \hat{p}_{\mathrm{E}}(\eta|s,a,\boldsymbol{\theta})$,

$$\boldsymbol{F}_{\hat{p}_{\mathrm{E}}}(s,a,\boldsymbol{\theta}) \triangleq -\int_\eta \hat{p}_{\mathrm{E}}(\eta|s,a,\boldsymbol{\theta}) \frac{\partial^2}{\partial \boldsymbol{\theta}^2} \log \hat{p}_{\mathrm{E}}(\eta|s,a,\boldsymbol{\theta})\, d\eta,$$

can be used as the Riemannian metric, because the FIM is a unique metric matrix of the second-order Taylor expansion of the KL divergence, *i.e.*, $D_{\mathrm{KL}}\bigl[\hat{p}_{\mathrm{E}}(\eta|s,a,\boldsymbol{\theta}),\hat{p}_{\mathrm{E}}(\eta|s,a,\boldsymbol{\theta}+\Delta\boldsymbol{\theta})\bigr] \simeq \frac{1}{2}\|\Delta\boldsymbol{\theta}\|^2_{\boldsymbol{F}_{\hat{p}_{\mathrm{E}}}(s,a)}$.

The steepest descent direction of a function $f(\boldsymbol{\theta})$ on a Riemannian space with a metric matrix $\boldsymbol{F}(\boldsymbol{\theta})$ is given by $-\boldsymbol{F}(\boldsymbol{\theta})^{-1}\partial f(\boldsymbol{\theta})/\partial\boldsymbol{\theta}$, which is called the natural gradient (NG) of $f(\boldsymbol{\theta})$ (Amari, 1998). Accordingly, to efficiently (locally) minimize $D_{\mathrm{KL}}$ with the NG, $\boldsymbol{\theta}$ is incrementally updated by

$$\boldsymbol{\theta} := \boldsymbol{\theta} + \alpha \boldsymbol{F}_{\hat{p}_{\mathrm{E}}}(s,a,\boldsymbol{\theta})^{-1} \tag{8}$$
$$\times \int_\eta \Pi^\pi \hat{p}_{\mathrm{E}}(\eta|s,a,\boldsymbol{\theta}) \frac{\partial}{\partial \boldsymbol{\theta}} \log \hat{p}_{\mathrm{E}}(\eta|s,a,\boldsymbol{\theta}) \mathrm{d}\eta.$$

The NG has another desirable property of parametric invariance (Amari and Nagaoka, 2000; Bagnell and Schneider, 2003) and thus the NG learning does not depend on the types of the parametrizations of the return density models. Also, if the approximated model with $\boldsymbol{\theta}$ includes the target model, the FIM becomes close to the Hessian matrix of the KL divergence near the optimal parameter $\boldsymbol{\theta}^*$ and the NG learning produces the Fisher-efficient online learning algorithm in the sense of asymptotic statistics (Amari, 1998). The FIMs of the distributions given in Eqs. (4), (5), and (6) can be computed analytically.

### 4.4 Stochastic natural gradient with samples

In most settings of RL, the transition of the state is stochastic. However, the gradient $\frac{\partial}{\partial \boldsymbol{\theta}} D_{\mathrm{KL}}$ of Eq. (7) cannot be computed directly, except for the case that state transitions and reward observations are deterministic. This is because the density $\Pi^\pi \hat{p}_{\mathrm{E}}$ given by Eq. (2), which appears in $\frac{\partial}{\partial \boldsymbol{\theta}} D_{\mathrm{KL}}$, includes the integrals about the state transition probability $p_{\mathrm{T}}(s_{+1}|s,a)$ and the reward density $p_{\mathrm{R}}(r|s,a,s_{+1})$ on $s_{+1}$ and $r$, whereas the learning agent observes only an actual successor state $s_{+1}$ and reward $r$. Therefore, $\Pi^\pi \hat{p}_{\mathrm{E}}$ is practically unknown. However, since the observed samples $s_{+1}$ and $r$ are drawn from $p_{\mathrm{T}}(s_{+1}|s,a)$ and $p_{\mathrm{R}}(r|s,a,s_{+1})$, respectively, $\Pi^\pi \hat{p}_{\mathrm{E}}$ can be estimated with the observed samples. Here we use the stochastic descent approach (Nocedal and Wright, 2006) as an implementation of the update of the return density estimate $\hat{p}_{\mathrm{E}}$, where the density $\Pi^\pi \hat{p}_{\mathrm{E}}$ in the deterministic (natural) gradient descent of Eq. (8) is replaced with a density $\tilde{p}_{\mathrm{E}}$ computed from the observed samples, such as

$$\boldsymbol{\theta} := \boldsymbol{\theta} + \alpha \boldsymbol{F}_{\hat{p}_{\mathrm{E}}}(s,a,\boldsymbol{\theta})^{-1} \tag{9}$$
$$\times \int_\eta \tilde{p}_{\mathrm{E}}\Bigl(\frac{\eta-r}{\gamma}\,|\,s_{+1},\boldsymbol{\theta}\Bigr) \frac{\partial}{\partial \boldsymbol{\theta}} \log \hat{p}_{\mathrm{E}}(\eta\,|\,s,a,\boldsymbol{\theta}) \mathrm{d}\eta.$$

There are two options for $\tilde{p}_{\mathrm{E}}$ in Eq. (9), being linked to the approximations for the Q function (Sutton and Barto, 1998).

(I) A Q-learning-type off-policy approach:

$$\tilde{p}_{\mathrm{E}}(\eta|s_{t+1},\boldsymbol{\theta}) := \hat{p}_{\mathrm{E}}(\eta|s_{t+1},\tilde{a}_{+1},\boldsymbol{\theta}), \tag{10}$$

where $\tilde{a}_{+1} := \underset{a_{+1} \in \mathcal{A}}{\mathrm{argmax}}\, \pi(a_{+1}|s_{+1})$.

(II) A SARSA-type on-policy approach:

$$\tilde{p}_{\mathrm{E}}(\eta|s_{t+1},\boldsymbol{\theta}) := \sum_{a_{+1} \in \mathcal{A}} \pi(a_{+1}|s_{+1}) \hat{p}_{\mathrm{E}}(\eta|s_{t+1},a_{+1},\boldsymbol{\theta}). \tag{11}$$

Below, we suppose use of a lookup table to represent the model parameters at each state-action pair. However, our approach can be extended to continuous state space with function approximators by the chain rule. For the Q-learning-type approach with the parametric density models defined in Eqs. (4), (5), and (6), the update rules with the NGs are given as follows[3].

(i) Update for the Gaussian model with the notation $\mu \triangleq \mu_{\boldsymbol{\theta}}(s,a)$, $\sigma \triangleq \sigma_{\boldsymbol{\theta}}(s,a)$, $\mu' \triangleq \mu_{\boldsymbol{\theta}}(s_{+1},\tilde{a}_{+1})$, $\sigma' \triangleq \sigma_{\boldsymbol{\theta}}(s_{+1},\tilde{a}_{+1})$, and a temporal difference $\delta \triangleq r + \gamma\mu' - \mu$:

$$\mu := \mu + \frac{\alpha}{\gamma}\delta, \tag{12}$$
$$\sigma := \sigma + \frac{\alpha}{\gamma}\{\delta^2 + \gamma^2\sigma'^2 - \sigma^2\}/2\sigma.$$

---

[3]Although the SARSA-type update rules can also be derived with Eq. (11), we omit these due to lack of space.

(ii) Update for the Laplace model with the notation $m \triangleq m_{\boldsymbol{\theta}}(s,a)$, $b \triangleq b_{\boldsymbol{\theta}}(s,a)$, $m' \triangleq m_{\boldsymbol{\theta}}(s_{+1}, \tilde{a}_{+1})$, $b' \triangleq b_{\boldsymbol{\theta}}(s_{+1}, \tilde{a}_{+1})$, and a temporal difference $\delta \triangleq r + \gamma m' - m$:

$$m := \begin{cases} m + \frac{\alpha}{\gamma}\{-1 + \exp(\frac{1}{\gamma b'}\delta)\}b, & \text{for } \delta \leq 0, \\ m + \frac{\alpha}{\gamma}\{1 - \exp(-\frac{1}{\gamma b'}\delta)\}b, & \text{for } \delta > 0, \end{cases} \quad (13)$$

$$b := b + \frac{\alpha}{\gamma}\left\{-b + |\delta| + \gamma b' \exp(-\frac{1}{\gamma b'}|\delta|)\right\}/2.$$

(iii) Update for the skewed Laplace model with the notation $m \triangleq m_{\boldsymbol{\theta}}(s,a)$, $b \triangleq b_{\boldsymbol{\theta}}(s,a)$, $c \triangleq c_{\boldsymbol{\theta}}(s,a)$, $m' \triangleq m_{\boldsymbol{\theta}}(s_{+1}, \tilde{a}_{+1})$, $b' \triangleq b_{\boldsymbol{\theta}}(s_{+1}, \tilde{a}_{+1})$, $c' \triangleq c_{\boldsymbol{\theta}}(s_{+1}, \tilde{a}_{+1})$, and a temporal difference $\delta \triangleq r + \gamma m' - m$:

- For $\delta \leq 0$,
$$\begin{cases} m := m + \frac{\alpha}{2\gamma c}\Big[ -2b - \frac{\gamma b'(1-c)(1-2c')}{c'(1-c')} - (1-c)\delta \\ \qquad + \frac{1-c'}{1-c}\left\{2b + \frac{\gamma b'(1-2c)}{c'}\right\}\exp(\frac{c'}{\gamma b'}\delta)\Big] \\ b := b + \frac{\alpha}{2\gamma c}\Big[b(c-1) - \frac{\gamma b'(1-c)^2(1-2c')}{c'(1-c')} - (1-c)^2\delta \\ \qquad + \frac{1-c'}{1-c}\left\{b(1-2c) + \frac{\gamma b'(1-3c+3c^2)}{c'}\right\}\exp(\frac{c'}{\gamma b'}\delta)\Big] \\ c := c + \frac{\alpha}{2\gamma b}\Big[b(c-1) - \frac{\gamma b'(1-c)^2(1-2c')}{c'(1-c')} - (1-c)^2\delta \\ \qquad + (1-c')\left\{b + \frac{\gamma b'(1-2c)}{c'}\right\}\exp(\frac{c'}{\gamma b'}\delta)\Big] \end{cases}$$

- For $\delta > 0$,
$$\begin{cases} m := m + \frac{\alpha}{2\gamma(1-c)}\Big[2b - \frac{\gamma b'c(1-2c')}{c'(1-c')} - c\delta \\ \qquad - \frac{c'}{c}\left\{2b - \frac{\gamma b'(1-2c)}{1-c'}\right\}\exp(\frac{c'-1}{\gamma b'}\delta)\Big] \\ b := b + \frac{\alpha}{2\gamma(1-c)}\Big[-bc + \frac{\gamma b'c^2(1-2c')}{c'(1-c')} + c^2\delta \\ \qquad + \frac{c'}{c}\left\{b(2c-1) + \frac{\gamma b'(1-3c+3c^2)}{1-c'}\right\}\exp(\frac{c'-1}{\gamma b'}\delta)\Big] \\ c := c + \frac{\alpha}{2\gamma b}\Big[-bc - \frac{\gamma b'c^2(1-2c')}{c'(1-c')} - c^2\delta \\ \qquad + c'\left\{b + \frac{\gamma b'(1-2c)}{1-c'}\right\}\exp(\frac{c'-1}{\gamma b'}\delta)\Big] \end{cases}$$

Fig. 1 shows examples of the NGs, which are the second terms in the right-hand sides of the above updates (with the learning coefficient $\alpha/\gamma$ set to 1). It is confirmed from Eq. (12) and Fig. 1 (A) that the update rule of the mean parameter $\mu$ in the Gaussian model is equivalent to the TD-learning for the expected return. Therefore, the algorithm with the Gaussian model can be regarded as a natural extension of the conventional TD-learning to the return density.

In contrast, it is confirmed from Eq. (13) and Fig. 1 (B) that the update quantity of the central parameter $m$ in the Laplace model is bounded with the scale parameter, $b$ ($\times \alpha/\gamma$). This indicates that the algorithm with the Laplace model is robust and can work stably even with some outlier samples. In the skewed Laplace model, Fig. 1 (C1) and (C2) indicate that, when the skewness parameter $c$ is larger than 0.5, the influence of a positive temporal difference $\delta > 0$ on the NGs of all the parameters $m$, $s$, and $c$ tend to be stronger than them of $\delta \leq 0$, where the temporal difference is $\delta \triangleq r + \gamma m' - m$.

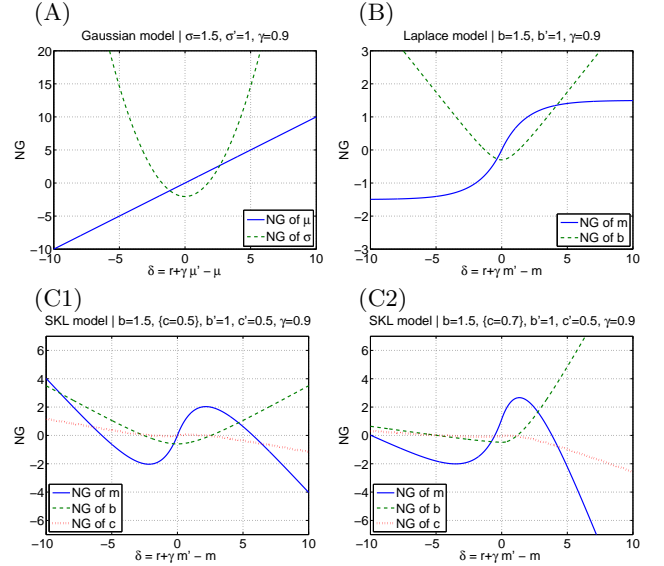

Figure 1: The examples of the NGs of the (A) Gaussian, (B) Laplace, and (C1,2) skewed Laplace distributions with respect to the temporal difference $\delta$.

## 5 Using the Estimated Return Density for Risk-Sensitive RL

Now we have a means to approximately evaluate the return densities. Then we can formulate RL algorithms with any criterion defined on the basis of a return density. As a demonstrative example, we show an approach using a quantile of the returns as the evaluation criterion for the action selection.

The $q$-quantile return conditioned on starting from a state-action pair with the quantile parameter $q \in [0,1]$, which we call a $q$-Q function, is defined as

$$Q_q^\pi(s,a) \triangleq \left\{ z \mid \int_{-\infty}^{z} p_{\mathrm{E}}^\pi(\eta|s,a)\, d\eta = q \right\}.$$

Given a $q$-Q function approximator $\hat{Q}_{q,\boldsymbol{\theta}}$ with parameters for a quantile $q$ and a return density model $\boldsymbol{\theta}$, we can heuristically define several policy model, such as the soft-max policy with the inverse temperature $\beta$:

$$\pi(a|s; q, \boldsymbol{\theta}, \beta) \triangleq \frac{\exp(\beta \hat{Q}_{q,\boldsymbol{\theta}}(s,a))}{\sum_{a \in \mathcal{A}} \exp(\beta \hat{Q}_{q,\boldsymbol{\theta}}(s,a))}. \quad (14)$$

The framework of the proposed $q$-quantile RL algorithms, which we call the $q$-Q learning and $q$-SARSA algorithms, is shown in Table 1. The quantity $\hat{Q}_{q,\boldsymbol{\theta}}$ is evaluated on the basis of the Gaussian, Laplace, or skewed Laplace distribution as the parametric family for the return density estimation as follows[4], where the same notations as the update rules in Section 4.4 are used.

---

[4]The conditional value at risk (Artzner et al., 1999) of the return is also derived analytically.

Table 1: Framework of the proposed algorithms

| The $q$-Q learning or $q$-SARSA Algorithm with NG |
|---|
| **Set:** $\pi$, $\hat{p}_{\mathrm{E}}$, $q \in (0,1)$, $\boldsymbol{\theta}$, $\alpha_t$, $s_0$ <br> **For** $t = 0$ **to** $T-1$ <br> $\quad a_t \sim \pi(a_t\|s_t; q, \boldsymbol{\theta})$ <br> $\quad s_{t+1} \sim p_{\mathrm{T}}(s_{t+1}\|s_t, a_t), \quad r_t \sim p_{\mathrm{R}}(r_t\|s_t, a_t, s_{t+1})$ <br> $\quad \boldsymbol{\theta} := \boldsymbol{\theta} + \frac{\alpha_t}{\gamma} \boldsymbol{F}_{\hat{p}_{\mathrm{E}}}(s_t, a_t, \boldsymbol{\theta})^{-1} \int_\eta \tilde{p}_{\mathrm{E}}(\frac{\eta - r_t}{\gamma}\|s_{t+1}, \boldsymbol{\theta})$ <br> $\quad\quad\quad\quad \times \frac{\partial}{\partial \boldsymbol{\theta}} \log \hat{p}_{\mathrm{E}}(\eta \| s_t, a_t, \boldsymbol{\theta}) \mathrm{d}\eta$ $\surd$ <br> **end** |

$\surd$ The definition of $\tilde{p}_{\mathrm{E}}$ is different in the $q$-Q learning or $q$-SARSA algorithm (see Eqs. (10) or (11)).

(i) In the Gaussian model: ('erf': Gauss error function)
$$\hat{Q}_{q,\boldsymbol{\theta}}(s,a) = \mu + \sigma\sqrt{2}\mathrm{erf}^{-1}(2q-1),$$

(ii) In the Laplace model:
$$\hat{Q}_{q,\boldsymbol{\theta}}(s,a) = \begin{cases} m + b\log(2q) & \text{for } q \leq 0.5, \\ m - b\log(2-2q) & \text{for } q > 0.5. \end{cases}$$

(iii) In the skewed Laplace model:
$$\hat{Q}_{q,\boldsymbol{\theta}}(s,a) = \begin{cases} m + \frac{b}{1-c}\log\frac{q}{c} & \text{for } q \leq c, \\ m - \frac{b}{c}\log\frac{1-q}{1-c} & \text{for } q > c. \end{cases}$$

The $q$-Q function can be regarded as a Value-at-Risk (VaR) for the return, which is widely used as the risk measurement in financial management (Artzner et al., 1999; Luenberger, 1998). An estimate of the VaR is often robust against outliers (Koenker, 2005). Consequently, this approach with a policy based on the $q$-Q function will provide instances of practical risk-sensitive and/or robust RL algorithms. However, it is known that the optimal policy maximizing any criterion (*e.g.*, $q$-quantile or mean-variance) other than the expected return is not necessarily a stationary policy in terms of the state (White, 1988; Defourny et al., 2008). Thus, in this paper, we focus only on a sub-optimal policy for the $q$-quantile criterion under the stationary policy class.

It is note that, as a similar work, Heger (1994) proposed a method of evaluating the 0- or 1-quantile return by extending the Bellman equation. However, no Bellman-type recursive formula holds with respect to the quantile at $q \in (0,1)$, and thus it is hard to directly evaluate the $q$-Q function with the conventional TD learning framework.

## 6 Numerical Experiments

We apply the proposed algorithms to a grid world task where the policy maximizing the expected return and the policy maximizing a $q$-quantile return are different. Various types of reward distributions are used to experimentally investigate the difference among the parametric models for the return densities.

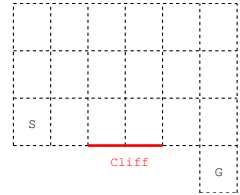

Figure 2: Cliff walk task

### 6.1 Task setting and results with deterministic reward

We used the well-known cliff walk task for benchmark purposes (Sutton and Barto, 1998). Our task involves an MDP with the infinite horizon (*i.e.*, non-episodic), 18 (= 6 × 3) states, and 4 actions. The map of the states (grids) is shown in Fig. 2, where the start and goal states are marked by 'S' and 'G', respectively. When an agent moves to the goal state, it is jumped to the start state. The four choices of actions correspond to moving from the current state to the north, south, east and west neighbor states, respectively. The agent goes in the chosen direction with probability 0.7 and goes in each of the other three directions with probability 0.1. If the agent attempts to leave from the grid world, the state will remain unchanged.

The setting of the reward function is as follows. If the agent arrives at the goal state, a deterministic positive reward $r_{\mathrm{goal}} = 12$ will be observed. If the agent walks toward the cliff (see Fig. 2), it will fall off the cliff and observe a negative reward $r_{\mathrm{cliff}} = -10$, and then the agent will return to the original state, *i.e.*, the successor state will be unchanged from the current state. Otherwise, the reward is always zero. This reward setting gives a situation in which the policy maximizing the expected return and the (stationary) policy maximizing the $q$-quantile return with small $q$ are different due to the risk of falling off the cliff.

This situation can be confirmed from Fig. 3, where the softmax policy (Eq. (14)) with time-varying $\beta_t$ was used. $\gamma$ and $\alpha$ were set as 0.95 and 0.1, respectively. The trajectories in the grid worlds in the upper side of Fig. 3 show the most probable state transitions of the policies obtained by Watkins' Q learning (Watkins and Dayan, 1992) and the $q$-Q learning. The curves plotted in the lower side of Fig. 3 are the time courses of $r_t$, $\beta_t$, and the expected and the 0.1-quantile returns from 'S'. These results indicate that the policy obtained by the $q$-Q learning could successfully avert the risk of falling off the cliff, while the policy by Watkins' Q learning seemed to take this risk.

Although we also applied the $q$-Q learning algorithms with the ordinary gradients (not with the NGs), those learning processes were much less stable compared with those with the NGs. So, we omit the details.

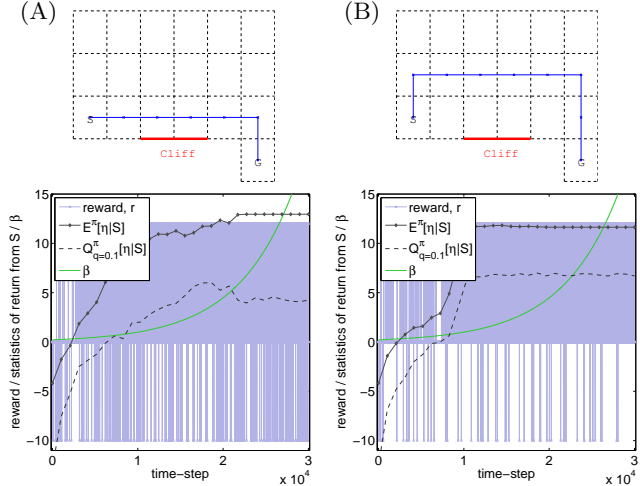

Figure 3: Typical learning results: (A) Q learning and (B) $q\!=\!0.1$-Q learning with the Gaussian distribution.

### 6.2 Results with various reward densities

Properties of the proposed algorithm may depend on the choice of the parametric models of return densities defined in Eqs. (4), (5), and (6). To highlight the dependence, we performed another set of numerical experiments comparing performance of these there choices, where the penalty of falling off the cliff, $r_{\text{cliff}}$, is assumed to follow either of the following three distributions with the common mean $\mathbb{E}[r_{\text{cliff}}] = -10$: (A) the (negative) gamma distribution with the shape and scale parameters set to 0.5 and 20, (B) Student's $t$-distribution with the degree of freedom and scale parameters set to 1.2 and 10, respectively. The gamma distribution is often used as the model of the losses in finance (Gordy, 2000). Thus it is important that the proposed algorithms for risk aversion are tested with rewards generated from the gamma distribution. On the other hand, a noise generated from Student's $t$-distribution with the degree of freedom $d \in (1, 2)$ has mean zero but infinite variance, and is useful for testing the robustness of the algorithms. Here, the $q \in \{0.1, 0.3, 0.5\}$-Q learning algorithms were tested. We also tested Watkins' Q-learning and the Q-hat learning (Heger, 1994) as the baseline algorithms. The total time-steps of the simulation $T$ was set to $3 \times 10^5$. The learning rate for learning the return density model parameter $\boldsymbol{\theta}$ was set to $\alpha_t = 1/(30 + 30t/T)$. The $\varepsilon$-greedy selection method is used for the policy (Sutton and Barto, 1998), where $\varepsilon_t$ was set as $\varepsilon_t = 1 - t/T$.

After the each simulation (trial), the averages and standard deviations of the following statistics were evaluated: the mean and $\{0.01, 0.1, 0.3, 0.5\}$-quantiles of the returns conditioned on starting from the start state 'S'. These were numerically computed by the Monte Carlo method with the learned policy and are presented in Table 2. If the results are significantly better or worse than Watkins' Q learning based on the t-test at the significance level 1%, the entries of the table will be indicated in the bold or italics, respectively. Note that, though the '$q\!:=\!x$'-Q learning seeks to maximize the '$x$'-quantile returns, the other statistics are also shown in Table 2.

Table 2 (A) shows the results when $-r_{\text{cliff}}$ was generated from the gamma distribution. The $q$-Q learning with the skewed Laplace model increased the $q$-quantile returns of small $q$ at most and could avert the risk. However, there is a tendency that the $q$-Q learning of $q:=x$ with the skewed Laplace model increased the $x'$-quantile return of $x' < x$ better than the the learning of $q:=x'$. This observation indicates that the $q$-Q learning with the skewed Laplace model tends to overestimate the risk of the returns. In contrast, the Q-hat learning for maximizing the worst-case did not work well even in the 0.01-quantile-return criterion.

Table 2 (B) shows the results when $r_{\text{cliff}}$ was generated from Student's $t$-distribution with infinite variance, where the $q$-Q learning with the Laplace model worked most appropriately. Especially in terms of maximizing the median ($q = 0.5$-quantile) returns, the $q$-Q learning with the Laplace model always worked stably against both the settings of $r_{\text{cliff}}$. Furthermore, even comparing with Watkins' Q learning with respect to the expected return in both the $r_{\text{cliff}}$ settings, the $q$-Q learning with the Laplace model worked better. This result indicates that the $q$-Q learning with the Laplace model could be robust, which is consistent with the theoretical discussion in Section 4.4.

However, these results also indicate that we will need to choose an appropriate density model for the return density estimation, depending on the task. Thus, model selection or combination will be significant.

### 7 Conclusion

We proposed a parametric approach for estimating return densities, which allows us to handle various criteria in a unified manner. A Bellman-type recursive formula for the return density was derived. We presented return-density-estimation algorithms with several parametric models by extending the TD-learning. The numerical experiments indicated that the proposed algorithms are promising in risk-sensitive and robust RL scenarios.

Further analysis of the proposed algorithms especially in terms of their convergences will be necessary to more deeply understand the properties and efficiency of our proposed approach to estimating the return densities. Also, comparison with an alternative approach to non-parametrically estimating return densities (Morimura et al., 2010) is important for our future work.

Table 2: Evaluations on the cliff walk MDP task.

(A) Penalty for the cliff is the random variable drawn from the negative gamma distribution.

| Algorithm | | Averages ± STDs of the statistics of the returns from the start state | | | | |
|---|---|---|---|---|---|---|
| Type | $q$ | Mean | 0.01-quantile | 0.1-quantile | 0.3-quantile | 0.5-quantile |
| Q learning | − | 12.75±0.58 | −13.58±9.47 | 6.20±1.84 | 11.27±0.58 | 13.77±1.17 |
| Q-hat learning | − | −0.09±0.07 | **−3.08**±1.58 | −0.00±0.02 | 0±0 | 0±0 |
| $q$-Q learning | 0.1 | 10.00±0.56 | **2.50**±1.59 | 6.58±0.55 | 8.76±0.55 | 10.19±0.56 |
| with $p^{\mathrm{g}}(\eta|s,a,\boldsymbol{\theta}^{\mathrm{g}})$ | 0.3 | 11.91±0.80 | **−2.09**±1.76 | 7.61±0.70 | 10.44±0.78 | 12.25±0.84 |
| $q$-Q learning | 0.1 | **13.06**±0.39 | −19.56±8.82 | 5.00±1.66 | **11.56**±0.40 | **14.53**±0.98 |
| with $p^{\mathrm{l}}(\eta|s,a,\boldsymbol{\theta}^{\mathrm{l}})$ | 0.3 | **13.26**±0.05 | −24.61±1.26 | 4.10±0.33 | **11.76**±0.12 | **15.05**±0.08 |
| | 0.5 | **13.26**±0.06 | −24.50±1.31 | 4.02±0.30 | **11.74**±0.11 | **15.03**±0.09 |
| $q$-Q learning | 0.1 | 10.84±1.25 | **0.35**±2.81 | **7.03**±0.99 | 9.48±1.22 | 11.07±1.34 |
| with $p^{\mathrm{skl}}(\eta|s,a,\boldsymbol{\theta}^{\mathrm{skl}})$ | 0.3 | 12.53±0.25 | **−6.10**±2.26 | **7.90**±0.18 | 11.07±0.23 | 13.03±0.30 |
| | 0.5 | **13.26**±0.05 | −24.69±1.40 | 4.09±0.27 | **11.73**±0.11 | **15.05**±0.08 |

(B) Penalty for the cliff is the random variable drawn from Student's $t$-distribution.

| Algorithm | | Averages ± STDs of the statistics of the returns from the start state | | | | |
|---|---|---|---|---|---|---|
| Type | $q$ | Mean | 0.01-quantile | 0.1-quantile | 0.3-quantile | 0.5-quantile |
| Q learning | − | 7.57±7.68 | −39.60±72.86 | −0.47±18.01 | 5.75±9.10 | 8.59±6.25 |
| Q-hat learning | − | −0.18±0.50 | **−7.47**±16.51 | −0.20±1.39 | 0±0 | 0±0 |
| $q$-Q learning | 0.1 | 6.74±4.04 | **−1.96**±7.76 | 4.02±2.90 | 5.62±3.79 | 6.77±4.27 |
| with $p^{\mathrm{g}}(\eta|s,a,\boldsymbol{\theta}^{\mathrm{g}})$ | 0.3 | 8.46±3.16 | **−2.45**±15.26 | 5.02±3.46 | 7.32±2.77 | 8.70±3.07 |
| $q$-Q learning | 0.1 | **11.94**±0.46 | −9.27±4.21 | **7.32**±0.75 | **10.44**±0.41 | **12.34**±0.37 |
| with $p^{\mathrm{l}}(\eta|s,a,\boldsymbol{\theta}^{\mathrm{l}})$ | 0.3 | **12.13**±0.54 | −17.52±15.63 | **6.63**±2.09 | **10.60**±0.44 | **12.73**±0.76 |
| | 0.5 | **13.25**±0.05 | −74.51±4.89 | −1.28±0.49 | **10.89**±0.14 | **15.10**±0.10 |
| $q$-Q learning | 0.1 | 9.99±1.25 | **−2.07**±3.77 | 6.27±1.14 | 8.68±1.28 | 10.23±1.32 |
| with $p^{\mathrm{skl}}(\eta|s,a,\boldsymbol{\theta}^{\mathrm{skl}})$ | 0.3 | 8.68±4.28 | −16.26±17.93 | 3.73±5.04 | 7.23±4.06 | 9.12±4.27 |
| | 0.5 | 8.38±3.55 | −29.87±22.50 | 1.79±4.13 | 6.67±3.37 | 9.08±3.75 |

## Acknowledgments

T.M. was supported by the PREDICT of the Ministry of Internal Affairs and Communications, Japan. M.S., H.K., and H.H. were supported by the FIRST program.